\documentclass{svproc}

\usepackage{url}

\newcommand{\hl}[1]{#1}

\usepackage{booktabs}
\usepackage{amssymb,amsmath,array}
\usepackage{graphicx} %
\usepackage{tikz}
\usepackage{algorithm}     %
\usepackage{algpseudocode} %
\usepackage{enumitem}
\usepackage{pgfplots}
\usepackage{pgf}

\pgfplotsset{compat=newest}

\usetikzlibrary{decorations.text, arrows.meta, positioning}
\usetikzlibrary{decorations.pathreplacing,calc,shapes,positioning}
\usetikzlibrary{shapes,shadows,arrows,positioning,angles,quotes}
\tikzset{My Arrow Style/.style={single arrow, fill=black!15, anchor=base, align=center,text width=2.3cm}}
\tikzstyle{arrow} = [thick,->,>=stealth]
\usetikzlibrary{calc,trees,positioning,arrows,fit,shapes,calc}

\tikzstyle{startstop} = [rectangle, rounded corners, minimum width=1.5cm, minimum height=0.5cm,text centered, draw=black, fill=red!30]
\tikzstyle{io} = [trapezium, trapezium left angle=70, trapezium right angle=110, minimum width=0.5cm, minimum height=0.5cm, text centered, draw=black, fill=blue!30]

\tikzstyle{process} = [rectangle, minimum width=3cm, minimum height=0.5cm, text centered, draw=black, fill=orange!30]
\tikzstyle{decision} = [diamond, minimum width=0.5cm, minimum height=0.1cm, text centered, draw=black, fill=green!30]
\tikzstyle{process2} = [rectangle, minimum width=1cm, minimum height=0.5cm, text centered, draw=black, fill=orange!30]
\tikzstyle{arrow} = [thick,->,>=stealth]

\usetikzlibrary{automata, positioning}
\usetikzlibrary{shapes.geometric, arrows, positioning}

\tikzstyle{block} = [rectangle, draw, text centered, rounded corners, minimum height=2em]
\tikzstyle{line} = [draw, -stealth, thick]
\tikzstyle{cloud} = [ellipse, draw, text centered, minimum height=2em, thick]
\tikzstyle{dashedcloud} = [ellipse, draw, dashed, text centered, minimum height=2em, thick]
\usetikzlibrary{positioning, arrows.meta, shapes.geometric, decorations.pathreplacing}

\usetikzlibrary{shapes.geometric, arrows.meta, positioning}

\begin{document}
\mainmatter

\title{Semi-Supervised CAPP Transformer Learning via Pseudo-Labeling\thanks{This work is funded by the European Union under grant agreement number 101091783 (MARS Project).}}
\titlerunning{Semi-supervised CAPP Transformer Learning}

\author{Dennis Gross\inst{1} \and Helge Spieker\inst{1} \and Arnaud Gotlieb\inst{1} \and Emmanuel Stathatos\inst{2} \and Panorios Benardos\inst{2} \and George-Christopher Vosniakos\inst{2}}

\authorrunning{Gross et al.} 
\institute{Simula Research Laboratory, Oslo Norway
\and
National Technical University of Athens, School of Mechanical Engineering, Manufacturing Technology Laboratory, Greece}

\maketitle

\begin{abstract}
\emph{High-level Computer-Aided Process Planning (CAPP)} maps part specifications to manufacturing process plans, but transformer-based CAPP models generalize poorly when labeled data is scarce, as it often is in industry.
We propose a semi-supervised approach that improves such models without additional manual labeling: a learned verifier (``oracle''), trained on the transformer's own generation statistics, filters the model's predictions on unseen parts, and only the sequences it judges correct are added to the training set for a single retraining cycle.
We evaluate on a fully enumerated synthetic CAPP benchmark across four low-data regimes (1--10\% of the space) over five seeds, comparing against no augmentation and unfiltered random augmentation.
Oracle-guided augmentation improves exact sequence accuracy by up to 11 percentage points in the 1\% regime, whereas unfiltered augmentation does not improve over the baseline, showing that the gains stem from selective curation rather than added data volume.
The method offers a practical way to bootstrap CAPP transformers in data-scarce manufacturing settings.
\keywords{Computer-Aided Process Planning, High-level CAPP, Semi-Supervised Learning, Machine Learning}
\end{abstract}

\section{Introduction}
\emph{High-level computer-aided process planning (CAPP)} sequences manufacturing operations based on part design specifications~\cite{elmaraghy1993evolution}, translating features such as geometry, surface finish, and tolerances into efficient process plans tailored to production requirements.
For example, a small-batch prototype with freeform geometry, coarse finish, and standard tolerances might be produced by sand casting for the near-net shape, followed by 5-axis milling for final accuracy. 
This is distinct from low-level CAPP, which focuses on predicting specific process parameters, e.g., milling speed, for each step in the process chain.

Traditionally, engineers generated such plans using expert knowledge and rule-based systems~\cite{wakhare2016sequencing}, which demand deep expertise and scale poorly across production contexts~\cite{rauch2015sustainability}.

Recent work addresses these limitations by framing high-level CAPP as a sequence prediction task~\cite{capp_gpt,azab2024capp} and applying \emph{transformer models}~\cite{DBLP:conf/nips/VaswaniSPUJGKP17,capp_gpt}. 
Instead of manually encoding rules, transformers learn from paired examples of part features and expert-generated process plans. 
Trained autoregressively, they generate complete plans token by token, where each token represents a single operation or feature in the sequence.

However, when training data is scarce, transformer performance degrades sharply~\cite{capp_gpt}.
To address this, we propose a \emph{semi-supervised learning approach}~\cite{mattera2024semi}.
\hl{A learned verifier, which we refer to as an oracle as it has no formal guarantees, is trained on available labeled data and aligned with the transformer's behavior to evaluate new predictions on unseen parts.}
We add only sequences judged correct to the augmented training set, then retrain the transformer.
This selective pseudo-labeling~\cite{DBLP:conf/cvpr/Ma24,DBLP:conf/nips/GoyalSRK22} improves generalization.
In our experiments, a complete CAPP dataset lets us simulate and measure the true impact of our semi-supervised pseudo-labeling method.
Although such datasets are rare in practice, the improvements demonstrate that semi-supervised CAPP transformer training via pseudo-labeling can benefit CAPP transformers in real-world settings.

Our approach has two key advantages: 
(1) no manual effort at deployment time, since the system processes new data automatically, and 
(2) selective augmentation, which reduces the risk of label corruption, making it practical for data-scarce industrial pipelines.

\textbf{Related Work.} 
The transformer architecture~\cite{DBLP:conf/nips/VaswaniSPUJGKP17} replaced recurrence~\cite{DBLP:conf/nips/SutskeverVL14} with self-attention, enabling efficient modeling of long-range dependencies.
This advance underpins recent transformer-based approaches in manufacturing~\cite{azab2024capp,ni2025large,capp_gpt,xu2024generative,holland2024large}.
A challenge in manufacturing is data scarcity~\cite{capp_gpt,papacharalampopoulos2024learning}. 
To reduce dependence on labeled data, semi-supervised learning techniques, such as pseudo-labeling~\cite{DBLP:conf/cvpr/XieLHL20,DBLP:conf/cfap/GrandvaletB05}, incorporate high-confidence model predictions into the training set.
\hl{These approaches are a form of bootstrapping, iteratively expanding a labeled set from the model's own predictions or adversarial inputs}~\cite{ouali2020overview}.
Variants refine pseudo-labels with confidence filtering, augmentation, or reasoning validation~\cite{DBLP:conf/cvpr/Ma24,DBLP:conf/nips/GoyalSRK22}.
These semi-supervised methods mitigate error reinforcement and improve generalization.
Instead of relying solely on model confidence, we follow work in weak supervision and noisy-label learning, where auxiliary oracles guide data selection~\cite{DBLP:journals/vldb/RatnerBEFWR20,DBLP:conf/aaai/Chakraborty20,DBLP:conf/wacv/PatelAQ23,jin2021evolutionary}.
Our approach trains a dedicated oracle on labeled data and on the trained CAPP transformer's behavior to validate test-time predictions and potentially improve retraining.
\hl{Unlike confidence-based self-training, which thresholds the model's own output probabilities, our novelty is a dedicated classifier trained on the transformer's internal generation statistics (logit- and entropy-derived features) to gate which pseudo-labels enter retraining. This decouples the accept/reject decision from the generator's miscalibrated confidence.}

\section{Methodology}
Our methodology transforms test-time predictions into training data using a learned oracle, improving a GPT-2-style CAPP transformer~\cite{capp_gpt} in low-resource settings without manual labeling.
We train the transformer on labeled data and fit a binary oracle on its errors. This oracle is then deployed to filter predictions on unseen parts; sequences classified as likely correct are added to the training set to retrain the model once and compare it against the baseline and random augmentation.
Simultaneously, those predictions classified as incorrect can be set aside for manual inspection and auditing by human experts to monitor weaknesses in model performance.

\subsection{High-Level CAPP Transformer}
\begin{figure}[t]
  \centering
  \resizebox{0.74\textwidth}{!}{%
  \begin{tikzpicture}[
      boxOrange/.style={rectangle, rounded corners=4pt, draw=orange!80!black, fill=orange!30, minimum width=3cm, minimum height=1cm, text centered, font=\normalsize},
      boxPurple/.style={rectangle, rounded corners=4pt, draw=purple!80!black, fill=purple!30, minimum width=3cm, minimum height=1cm, text centered, font=\normalsize},
      boxBlue/.style={rectangle, rounded corners=4pt, draw=blue!80!black, fill=blue!30, minimum width=3cm, minimum height=1cm, text centered, font=\normalsize},
      boxGreen/.style={rectangle, rounded corners=4pt, draw=green!80!black, fill=green!30, minimum width=3cm, minimum height=1cm, text centered, font=\normalsize},
      decision/.style={diamond, draw=black!70, fill=gray!10, aspect=2.5, text width=2.2cm, align=center, inner sep=0pt, font=\normalsize},
      arrow/.style={-{Latex[length=3mm,width=2mm]}, line width=1pt}
  ]
  \node[boxOrange] (encoding) {Part Encoding};
  \node[boxPurple, right=1.4cm of encoding] (tokenized) {Tokenized Input};
  \node[decision, right=1.4cm of tokenized] (decide) {Is End of \\ Sequence?};
  \node[boxGreen, right=1.4cm of decide] (finalSeq) {Final Sequence};

  \node[boxOrange, above=1cm of encoding] (order) {Part Order};
  \node[boxPurple] (outToken) at (tokenized |- order) {Output Token};
  \node[boxBlue] (capp) at (decide |- order) {CAPP Transformer};
  \node[boxGreen] (viable) at (finalSeq |- order) {Viable Process Plans};

  \draw[arrow] (order.south) -- (encoding.north);
  \draw[arrow] (outToken.south) -- (tokenized.north);
  \draw[arrow] (finalSeq.north) -- (viable.south);

  \draw[arrow] (encoding.east) -- (tokenized.west);
  \draw[arrow] (tokenized.east) -- (decide.west);
  \draw[arrow] (decide.east) -- node[midway, above] {YES} (finalSeq.west);

  \draw[arrow] (decide.north) -- node[midway, right] {NO} (capp.south);
  \draw[arrow] (capp.west) -- (outToken.east);
  \end{tikzpicture}
  }
  \caption{Flowchart showing the process of generating viable process plans (adapting \cite{capp_gpt}).}
  \label{fig:flowchart}
\end{figure}
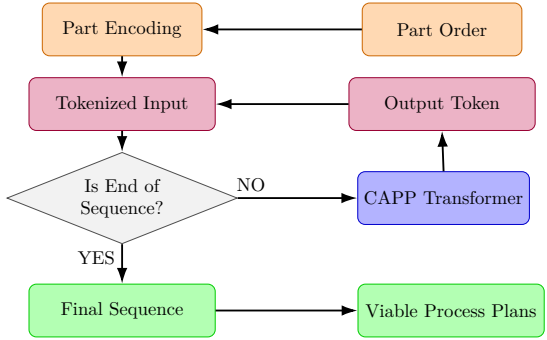

Following~\cite{capp_gpt}, we frame high-level CAPP as an autoregressive sequence generation task, implemented through a transformer model~\cite{DBLP:conf/nips/VaswaniSPUJGKP17}.
Transformers generate sequences autoregressively, predicting each token based on the input and already-predicted outputs until an end-of-sequence marker is reached. %
Training adjusts the model so that correct tokens receive higher probability, enabling it to learn patterns and produce coherent outputs.%
In high-level CAPP, the transformer takes a fixed part encoding and generates feasible process chains, which are ordered lists of manufacturing operations, token by token until completion~\cite{capp_gpt} (see Figure~\ref{fig:flowchart}). 
A single part encoding can have multiple feasible process chains as alternative ways to produce the part.

\subsection{Oracle}
To make test-time predictions usable as additional training data, we introduce a learned oracle that estimates whether a generated sequence is likely correct. 
Rather than relying solely on the transformer's own confidence, the oracle provides a different perspective, while being highly adapted to the transformer's data distribution, since it is trained on examples labeled from the transformer outputs.
We formulate the oracle as a binary classification problem to distinguish between correct and incorrect transformer outputs.

For each sequence generation, we collect step-wise logits ${L} = [{l}_1, {l}_2, \ldots, {l}_T] \in \mathbb{R}^{T \times |\Sigma|}$ and derive probability distributions ${P} = \text{softmax}({L})$. We extract a comprehensive high-dimensional feature vector ${z}$ capturing 132 features in total:
\emph{Confidence Patterns}: Distributional statistics of maximum, second-highest, and third-highest token probabilities across the sequence.
\emph{Uncertainty Quantification}: Entropy and perplexity measures characterizing model uncertainty at each timestep.
\emph{Decision Quality}: Probability margins between top-ranked token choices.
\emph{Distribution Properties}: Gini coefficients, KL divergence from uniform distribution, and variance measures capturing probability concentration.
\emph{Temporal Dynamics}: First-order differences, linear trends, and autocorrelation coefficients revealing how confidence and uncertainty evolve during generation.
\emph{Sequence-Level Patterns}: Token repetition rates, end-of-sequence behavior, cumulative measures, sequence length, and likelihood-based quality indicators. 
\emph{Padded sequence:} \hl{The per-timestep maximum probabilities, right-padded with zeros to the maximum sequence length so all feature vectors share a fixed dimensionality,} to capture temporal patterns.

As an oracle architecture, we use an XGBoost classifier~\cite{chen2016xgboost}. However, we do not see any limitation in choosing another classifier that assigns a class based on the likelihood of an error.
One alternative to our approach is simple confidence-thresholding, where sequences are accepted if the transformer's average sequence probability exceeds a threshold. 
However, in autoregressive CAPP sequence generation, a simple threshold can be misleading and insufficient. 
The transformer model might be highly confident in an incorrect sequence, or show low confidence at a single complex manufacturing step, but still output an entirely valid plan.
We therefore introduce a dedicated model that serves as a dynamic threshold function, which learns specific failure modes of the transformer.
For example, it can learn whether a sudden spike in entropy for one output token indicates a fatal sequence error or just a viable alternative branching path. 
Thus, the oracle utilizes the transformer's internal states, and learns to judge them accordingly.

\section{Experiments}
We evaluate our methodology across different training data regimes to test its effectiveness in improving CAPP transformer accuracies. 
Our transformer models are trained GPT-2 style decoder-only transformers from previous work~\cite{capp_gpt}. 

\textbf{Dataset and Evaluation Protocol.} 
For our experiments, we need to handle three distinct data splits: 
(1) a labeled training set for the transformer training $D_{train}$, which is additionally split into training and validation data; 
(2) a pool for pseudo-labeling $D_{pool}$; 
and (3) a strictly held-out test set $D_{test}$:

\begin{enumerate}[noitemsep,topsep=2pt]
    \item \textbf{Base Training}: The transformer is initially trained on $D_{train}$.
    \item \textbf{Oracle Training}: The oracle is trained on the transformer's outputs on $D_{train}$. \hl{Incorrect outputs form the positive (error) class and correct outputs the negative class. These classes are imbalanced (roughly even for \texttt{capp1}, with an error rate of $\approx0.48$, but skewed toward correct at higher label fractions, $\approx0.10$ for \texttt{capp10}). We apply inverse-frequency reweighting of the error class in XGBoost.}
    \item \textbf{Pseudo-Labeling}: The trained transformer generates predictions for the unseen parts in $D_{pool}$. The oracle scores these predictions. Sequences classified as correct are added to $D_{train}$ to form $D_{augmented}$.
    \item \hl{\textbf{Retraining}: The transformer is retrained with $D_{augmented}$}.
    \item \textbf{Evaluation}: The retrained transformer is evaluated exclusively on the held-out $D_{test}$. There is no intersection between the test data and the data used for training, pseudo-labeling and augmentation.
\end{enumerate}

The total dataset $D$, generated using a custom part encoding and rule-based algorithm~\cite{capp_gpt}, spans the entire input space (an uncommon case where all possible data points are available).
This makes common standard split ratios (e.g., 70\% training) unrealistic and thereby unsuitable.
To simulate limited-data conditions and assess generalization, we train models on progressively larger subsets of the dataset $D_{train}$: \texttt{capp1} (1\%), \texttt{capp2.5} (2.5\%), \texttt{capp5} (5\%), and \texttt{capp10} (10\%).
The oracle accuracy on the held-out $D_{test}$ is 80\% (\texttt{capp1}), 91.59\% (\texttt{capp2.5}), 97.19\% (\texttt{capp5}), and 99.54\% (\texttt{capp10})\hl{, with ROC-AUC ranging from 0.89 (\texttt{capp1}) to 0.94 (\texttt{capp10})}.

\textbf{Metric.} 
We evaluate the CAPP model performance using the exact \emph{Sequence Accuracy}. Sequence accuracy means that the entire process plan of all alternative process chains must be predicted entirely correctly, without missing or wrongly ordered process steps. This is the strictest, but at the same time most relevant metric, as any wrong process chain prediction can cause problems in the downstream process. 
For example, in a distributed manufacturing setting, a missing process step might cause the assignment of the part manufacturing to a factory that is missing the necessary machinery, or misordered process steps might cause defects in the produced parts.

\textbf{Execution.} 
At a high level, our experiment follows the following idea: start with an original transformer trained on the available labeled data from~\cite{capp_gpt}, then ask whether adding extra data (via different approaches) helps. 
We compare three independent approaches. 
First, a baseline transformer is \textit{retrained only on the original labeled set} (no data is added). 
Second, we apply \textit{our verifier oracle}, which filters model outputs and selects only those it detects as likely correct, and retrain on this augmented set. 
Third, we create a \textit{random augmentation baseline} by adding the same number of model-generated sequences, chosen randomly without filtering. 
We ran our experiments on a laptop with 16 GB RAM, an AMD Ryzen 7 7735HS under Ubuntu 20.04.5 LTS, and an NVIDIA GeForce RTX 3070 Ti Laptop GPU.

By holding the model architecture and training setup fixed and using the transformers and datasets from~\cite{capp_gpt}, and varying only how the additional data is chosen, we can directly see whether the oracle adds value beyond simply increasing dataset size.
The additional data is selected based on the proportion of model-generated sequences classified as correct by the oracle, with an equal number of randomly chosen sequences added for the random baseline.

\begin{table}[t]
\centering
\caption{Sequence accuracy (mean\,$\pm$\,std, \%) across 5 seeds on the fixed held-out test set.
Oracle$_N$ adds the top-$N$\% of oracle-approved sequences; Random$_N$ adds the same count drawn uniformly from the pool.
Best augmented results in \textbf{bold}.}
\label{tab:results_multiseed}
\scalebox{0.87}{%
\begin{tabular}{@{}lcc|cc|cc|cc@{}}
  \toprule
  Dataset
    & Original & Baseline
    & Oracle\textsubscript{10} & Random\textsubscript{10}
    & Oracle\textsubscript{50} & Random\textsubscript{50}
    & Oracle\textsubscript{100} & Random\textsubscript{100} \\
  \midrule
  \texttt{capp1}  
    & $49.1\pm0.0$ & $51.9\pm0.8$
    & $53.2\pm1.2$ & $51.6\pm2.2$
    & $56.9\pm1.4$ & $52.3\pm1.5$
    & $\mathbf{60.3\pm1.1}$ & $51.0\pm1.0$ \\
  \texttt{capp2.5} 
    & $87.8\pm0.0$ & $88.9\pm0.6$
    & $89.1\pm0.9$ & $89.2\pm0.7$
    & $89.7\pm0.7$ & $88.6\pm0.7$
    & $\mathbf{90.4\pm1.2}$ & $88.6\pm0.4$ \\
  \texttt{capp5}   
    & $97.4\pm0.0$ & $97.7\pm0.2$
    & $98.3\pm0.3$ & $98.3\pm0.3$
    & $98.2\pm0.3$ & $98.1\pm0.3$
    & $98.3\pm0.2$ & $\mathbf{98.6\pm0.1}$ \\
  \texttt{capp10}  
    & $99.3\pm0.0$ & $99.9\pm0.1$
    & $\mathbf{99.9\pm0.1}$ & $99.8\pm0.2$
    & $\mathbf{99.9\pm0.1}$ & $99.6\pm0.1$
    & $99.5\pm0.2$ & $99.7\pm0.1$ \\
  \bottomrule
\end{tabular}
}
\end{table}

\textbf{Results.} 
\hl{%
Table~\ref{tab:results_multiseed} reports mean $\pm$ std across 5 random seeds.
In the data-scarce regime, Oracle selection outperforms random augmentation:
in \texttt{capp1}, Oracle$_{100}$ gains $+11.2$\,pp over the original model and $+8.4$\,pp over the retrained baseline,
while Random$_{100}$ falls $0.9$\,pp \emph{below} baseline.
The Oracle--Random gap grows monotonically with augmentation level:
$+1.6$\,pp at 10\%, $+4.6$\,pp at 50\%, and $+9.3$\,pp at 100\% of oracle-approved sequences.
The gains therefore track curation rather than volume: adding unfiltered sequences does not improve over baseline, and at full scale Random$_{100}$ falls below it (though not significantly; see below).
This curation effect is visible directly in the accepted set: in \texttt{capp1}, oracle-accepted sequences are $95.6\pm2.6$\%, $96.1\pm0.7$\%, and $92.4\pm1.7$\% correct at 10\%, 50\%, and 100\% augmentation, versus only $50.2\pm1.9$\% for random selection, so oracle filtering keeps $D_{augmented}$ substantially cleaner.
In higher-data regimes (\texttt{capp5}, \texttt{capp10}), all conditions converge near ceiling and differences fall within run-to-run variability, consistent with saturation.
Paired $t$-tests over the five seeds are significant for the Oracle advantage in \texttt{capp1}:
Oracle$_{100}$ vs.\ Random$_{100}$ ($t=14.0$, $p<0.001$) and Oracle$_{50}$ vs.\ Random$_{50}$ ($t=5.5$, $p<0.01$).
In \texttt{capp2.5} the direction is the same but the smaller absolute differences give borderline significance ($p\approx0.06$ for Oracle$_{100}$ vs.\ Random$_{100}$).
The Random$_{100}$ drop below baseline in \texttt{capp1} is not significant across seeds ($p=0.25$): unfiltered augmentation is unreliable rather than reliably harmful.
The Oracle gains, by contrast, are consistent across the five seeds.}

\textbf{Discussion.}  
From a practical perspective, the approach integrates naturally into manufacturing pipelines where domain expertise is scarce, expensive, and fragmented across specialists and geographic locations.
Traditionally, building enough training data to deploy a CAPP transformer would require extensive expert planning and labeling. 
With oracle-guided pseudo-labeling, manufacturers can bootstrap CAPP models from a small initial dataset. The model then improves continuously as unannotated part specifications arrive. 
This has two advantages: The model adapts to the observed production data distribution, and the base dataset grows over time, supporting a later retraining of the CAPP transformer and oracle with an extended dataset, if needed.

Since no human intervention is required at deployment time, the framework scales and avoids costly expert annotations, creating an efficient human-in-the-loop workflow where the expert is no longer the bottleneck.
While the oracle automatically routes its approved (``likely correct'') sequences back into the training pipeline to improve the model, the sequences flagged as ``incorrect'' or highly uncertain can be set aside for manual review. 
Expert engineers then review only the edge cases, rather than planning every standard part from scratch.

This capability is relevant in industrial settings where the space of producible parts is vast.
We use a parametrically defined finite feature space to establish a reliable ground truth.
Nevertheless, the combinatorial explosion of part features (geometry, tolerances, batch sizes) mirrors the high-dimensional, near-infinite input spaces of real manufacturing. 
In these realistic scenarios, no initial training dataset can cover all possible part combinations. 
Data augmentation at test (i.e., deployment) time, helps precisely here: it expands the model's competence into input regions underrepresented in the original training data.

Nevertheless, limitations remain. The absolute accuracy of transformers in extremely low-resource settings (e.g., our 1\% setting) remains modest, even with augmentation. 
\hl{Similarly, an accepted-but-wrong pseudo-label can negatively influence model performance.
We deliberately use a single pseudo-labeling and retraining cycle rather than iterating. Repeated cycles risk confirmation bias and error reinforcement, where the model's own mistakes are recycled as training targets. This is a danger that grows in real deployments, where the oracle's accuracy cannot be independently verified against an enumerated ground truth as it can here.}
While pseudo-labeling can boost the performance, it is no substitute for representative training data.
\hl{All experiments moreover use a parametrically generated synthetic dataset, chosen because full enumeration is what makes the oracle's true accuracy measurable against known ground truth; validation on real part datasets remains future work.}
Deploying such systems in highly demanding, real-world manufacturing environments will still require complementary strategies, both to enhance the training data and to verify the correctness of the produced process chains.

\section{Conclusion}
We proposed a semi-supervised pseudo-labeling method that improves transformer-based CAPP models by retraining on unseen outputs estimated to be correct.
By extracting high-dimensional statistical features from the transformer's internal generation process, our trained oracle filters newly generated, unlabeled process plans, identifying correct sequences with high accuracy and no manual labeling.
Our experiments demonstrate that retraining the transformer on these autonomously verified predictions yields consistent sequence accuracy gains without any additional manual labeling effort. 
We found that the benefits are highest in heavily data-constrained environments, with sequence accuracy improvements by up to 11 percentage points in the lowest-data regime tested.

This approach is applicable to more complex systems, without requiring extensive datasets upfront, but instead supporting continuous data collection and model refinement. 
Future work will explore integrating specific part semantics and physical manufacturing constraints into the oracle's feature space, as well as evaluating the pipeline's performance in an active-learning setting where rejected plans are corrected by human experts.

\bibliographystyle{unsrt}
\bibliography{references}
\end{document}